\documentclass[times,preprint,12pt]{elsarticle}
\usepackage{amsmath, amssymb, amsfonts}
\usepackage{orcidlink}
\usepackage{algorithm2e}

\newtheorem{lemma}{Lemma}
\usepackage{booktabs}
\usepackage{makecell}
\usepackage{color,array}
\usepackage{graphicx}
\usepackage{comment}
\usepackage{pdflscape}
\usepackage{rotating} 
\usepackage{textcomp}
\usepackage[table,xcdraw]{xcolor}
\usepackage{subcaption}
\newcommand{\vh}[1]{\rotatebox{90}{\makecell[c]{#1}}}

\begin{document}

\begin{frontmatter}

\title{Interpretable Unsupervised Deformable Image Registration via Confidence-bound Multi-Hop Visual Reasoning} 

\author[inst1]{Zafar Iqbal}
\author[inst1]{Anwar Ul Haq}
\author[inst2]{Srimannarayana Grandhi}

\affiliation [inst1]
{organization={School of Engineering and Technology},
            addressline={Central Queensland University}, 
           city={Sydney},
           country={Australia}}

\affiliation [inst2]
{organization={School of Engineering and Technology},
            addressline={Central Queensland University}, 
            city={ Melbourne},
            country={Australia}}

\begin{abstract}
Unsupervised deformable image registration requires aligning complex anatomical structures without reference labels, making interpretability and reliability critical. Existing deep learning methods achieve considerable accuracy but often lack transparency, leading to error drift and reduced clinical trust. We propose a novel Multi-Hop Visual Chain of Reasoning (VCoR) framework that reformulates registration as a progressive reasoning process. Inspired by the iterative nature of clinical decision-making, each visual reasoning hop integrates a Localized Spatial Refinement (LSR) module to enrich feature representations and a Cross-Reference Attention (CRA) mechanism that leads the iterative refinement process, preserving anatomical consistency. This multi-hop strategy enables robust handling of large deformations and produces a transparent sequence of intermediate predictions with a theoretical bound. Beyond accuracy, our framework offers built-in interpretability by estimating uncertainty via the stability and convergence of deformation fields across hops. Extensive evaluations on two challenging public datasets, DIR-Lab 4D CT (lung) and IXI T1-weighted MRI (brain), demonstrate that VCoR achieves competitive registration accuracy while offering rich intermediate visualizations and confidence measures. By embedding an implicit visual reasoning paradigm, we present an interpretable, reliable, and clinically viable unsupervised medical image registration.
\end{abstract}


\end{frontmatter}

\section{Introduction}
\label{intro}

Medical image registration plays a central role in advancing clinical diagnostics by establishing precise spatial correspondence between images acquired under varying conditions across different modalities, time points, or subjects  \cite{ding2024c2fresmorph, iqbal2024hybrid, jiang2025tumor, jiang2025deformable}. When performed accurately, registration enables robust longitudinal tracking of disease progression, precise guidance of complex surgical  and interventional procedures, and the construction of high-fidelity anatomical atlases for population-level analysis \cite{zhang2025unsupervised, iqbal2025implicit, chen2022transmorph}. Reliable registration not only enhances visual interpretation but also transforms raw medical images into actionable knowledge that supports data-driven decision-making in healthcare \cite{zhang2025unsupervised}.

Deformable image registration (DIR) presents substantially greater complexity than rigid or affine registration, as it requires estimating dense, non-linear spatial transformations to accurately capture local anatomical variations and tissue deformations \cite{huang2026multi, iqbal2024hybrid, chen2025survey}. This complexity arises from the high dimensionality of the transformation space, the need to balance spatial accuracy with smoothness constraints, and sensitivity to modality differences, noise, and anatomical variability across subjects \cite{huang2026multi, wu2025cascaded}. A deformation vector field (DVF) $\mathbf{u}: \Omega \to \mathbb{R}^D$ maps each point $\mathbf{x}$ in the domain $\Omega$ of a source image $S_{im}$ to its corresponding location in a reference image $R_{im}$. The objective is to determine $\mathbf{u}$ such that the warped source image $S_{im,w}(\mathbf{x}) = S_{im}(\mathbf{x} + \mathbf{u}(\mathbf{x}))$ is optimally aligned with the reference image $R_{im}(\mathbf{x})$. This alignment is quantitatively assessed by maximizing image similarity metrics, such as the Dice Similarity Coefficient (DSC) for structural overlap and Normalized Cross-Correlation (NCC) for intensity pattern correspondence, while minimizing dissimilarity metrics such as Mean Squared Error (MSE), target registration error (TRE), and negative Jacobian (NJ) \cite{iqbal2024hybrid}. The transformation $\mathbf{T}(\mathbf{x}) = \mathbf{x} + \mathbf{u}(\mathbf{x})$ is inherently non-linear, enabling the capture of complex anatomical deformations. 

Traditional deformable registration methods employ iterative optimization to minimize a cost function that balances image similarity with a regularization term enforcing smoothness and physical plausibility of the deformation field \cite{cheng2025comprehensive, iqbal2024unsupervised, chen2021vit}. However, these approaches often fail to preserve anatomical consistency in complex or large-scale 3D medical datasets when no supervision is available, highlighting the need for unsupervised yet anatomically robust registration models \cite{zhang2022deformable, chen2025coarse, meng2025autofuse, chen2025dual}. While deep learning has achieved remarkable successes in medical image registration, the inherent “black-box” nature of these models \cite{hu2025improving, huang2025diffusion} raises critical concerns about interpretability and reliability in clinical settings. This lack of transparency and accountability remains a major barrier to adoption, underscoring the urgent need for models that are both accurate and trustworthy \cite{cheng2025comprehensive}.

To fill this research gap, we propose an unsupervised deformable image registration framework that achieves accurate and interpretable anatomical alignment through a multi-hop Visual Chain-of-Reasoning (VCoR). Our approach is inspired by cognitive studies of human decision-making, particularly Diffusion Decision Models \cite{khoudary2025reasoning, ratcliff2008diffusion}, which demonstrate that confidence increases as evidence accumulates while uncertainty correspondingly declines \cite{hellmann2024confidence, lu2025closing}. Translating this principle into deformable registration, we introduce a theoretically grounded confidence–consistency constraint that enforces monotonic growth in confidence and progressive reduction in uncertainty across successive reasoning steps. This yields an implicit VCoR process, where the deformation field is refined step by step, ensuring each hop produces more confident and anatomically plausible alignments than the last. To reinforce consistency, we employ a unidirectional cross-attention mechanism that progressively strengthens the correspondence of visual features between source and target images. By embedding interpretability within the registration process, our model delivers both high accuracy and transparent reasoning trails, fostering trust in clinical deployment.

 Our approach makes the following contributions:

\begin{enumerate}
    \item  The proposed method progressively improves the deformation vector field (DVF), with a coarse-to-fine, multi-hop refinement pipeline. To the best of our knowledge, this is the first work to introduce Implicit Multi-Hop Visual Chain-of-Reasoning (IMVCoR) for unsupervised deformable medical image registration.
    
    \item We propose a reliability bound, inspired by Diffusion Decision Model \cite{ratcliff2008diffusion}, embedding human-like reasoning into deep learning with empirical evidence. This mitigates error drift, preserves anatomical plausibility, and confirms validity through reduced negative Jacobians.
    
    \item We evaluate our method on DIR-Lab lung CT and IXI brain MRI benchmarks, achieving over 20\% TRE reduction, Dice improvements up to 5\%, and fewer negative Jacobians than state-of-the-art unsupervised approaches, validating both effectiveness and generalizability.

\end{enumerate}

\section{Related Work}

Deep learning (DL) has revolutionized medical image analysis, enabling significant gains in computational efficiency and accuracy across tasks such as image registration \cite{liu2025focusmorph, iqbal2024unsupervised, zhu2025symmetric}. Registration models, typically built on encoder–decoder architectures such as U-Nets \cite{ronneberger2015unet}, learn to estimate dense deformation vector fields (DVFs) directly from input image pairs \cite{dalca2019voxelmorph, cao2024light}. This paradigm shift moves the computational burden from inference to training, thereby enabling near real-time registration once the model has been trained \cite{srinivas2025robust}.

\begin{figure*}[htbp]
   \centering    \includegraphics[width=1 \linewidth, height=9cm]{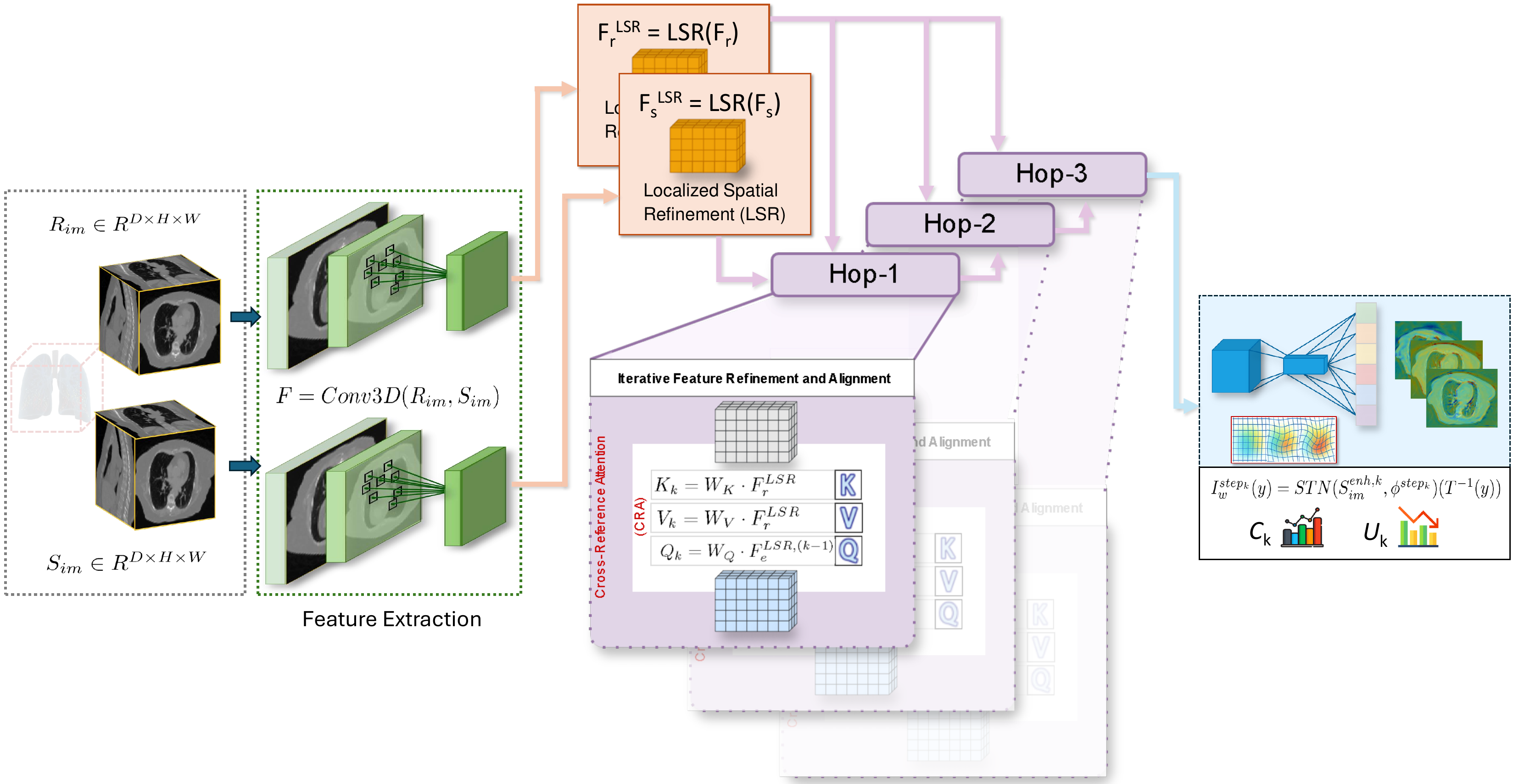}
   \caption{The proposed VCoR architecture performs three progressive reasoning hops, each combining a Localized Spatial Refinement (LSR) and Cross-Reference Attention (CRA) block. This design ensures anatomically consistent alignment while providing interpretability through hop-wise intermediate outputs.}
   \label{fig:VCoRmetho}
\end{figure*}

To overcome the limitations of supervised methods, unsupervised DL approaches have gained significant attention in deformable image registration (DIR). These methods avoid the need for ground-truth DVFs by formulating registration as an image reconstruction problem, where networks are trained to directly optimize similarity metrics between fixed and warped moving images \cite{balakrishnan2019voxelmorph, de2019deep, wang2025equivariance}. Recent work includes Gaussian-based models that explicitly estimate uncertainty, as well as robust frameworks designed to handle noisy data \cite{li2024gaussian}. Specialized architectures such as decoder-only designs have also shown improvements in accuracy and efficiency \cite{jia2025decoder}. More recently, Mamba architectures have emerged as efficient alternatives to transformers, offering enhanced long-range dependency modeling while preserving contextual information. Despite these advantages, the architectural complexity of Mamba may hinder interpretability and require careful optimization to ensure stable convergence across diverse imaging modalities \cite{wang2025mamba}.

Beyond unsupervised methods, research has increasingly focused on persistent challenges such as multimodal alignment, handling large deformations, and improving computational efficiency. Hybrid strategies that integrate Coarse-to-Fine Vision Transformers (ViTs) with Convolutional Neural Networks (CNNs) \cite{chen2025coarse} enhance robustness under significant anatomical variation, while dual-domain models that optimize phase consistency using Global Intersection Mutual Information (GIMI) achieve higher accuracy by better modeling inter-modal dependencies \cite{chen2025dual}. To further refine registration quality, resolution-enhancement plug-ins have been proposed as modular add-ons, preserving fine anatomical structures \cite{sun2023resolution}. Efficiency-driven designs such as large-kernel convolutions with rectangular decomposition \cite{cao2024light} reduce computational overhead while maintaining accuracy, supporting potential real-time applications. 
Attention-based innovations include the Equivariance and Motion-Enhanced Neighborhood Attention-based Pyramid (NAP) network \cite{wang2025equivariance}, wavelet-transform models combined with linear attention for multi-scale contextual reasoning \cite{li2024deformable}, and descriptor-based approaches that integrate dense features with Directed Edge Enhancers (DEE) for robust multimodal tasks such as COVID-19 image registration \cite{srinivas2025robust}. Collectively, these studies reflect a growing trend toward integrating attention mechanisms, transformer-based architectures, and efficient convolutional strategies to address the inherent complexities of modern medical image registration. 

\section{Proposed Methodology}
\label{sec:methodology}

Our proposed framework, illustrated in Figure.~\ref{fig:VCoRmetho}, addresses anatomically consistent and interpretable deformable medical image registration by progressively refining the alignment process through a Visual Chain of Reasoning (VCoR) approach. The methodology, detailed in Algorithm~\ref{alg:VCoR}, integrates advanced deep learning components, including 3D U-Net-based feature extraction, self-attention, and iterative cross-attention, to achieve accurate registration while providing greater transparency into the model’s reasoning process.

\textit{\textbf{Problem Description}}:  
Given a source image $S_{im}: \Omega \to \mathbb{R}$ and a reference image $R_{im}: \Omega \to \mathbb{R}$, the objective is to learn a mapping function $G: (S_{im}, R_{im}) \mapsto \mathbf{u}$ that predicts a deformation vector field (DVF) $\mathbf{u}: \Omega \to \mathbb{R}^D$. The DVF transforms the source image into a warped image  

\begin{equation}
S_{im,w}(\mathbf{x}) = S_{im}\big(\mathbf{x} + \mathbf{u}(\mathbf{x})\big),
\end{equation}

such that the warped image $S_{im,w}$ is accurately aligned with the reference image $R_{im}$.  

The overall architecture comprises three main stages: (1) a feature extraction module, (2) a multi-stage VCoR alignment module, and (3) a deformation field generation and image warping component. The VCoR module is central to this design, iteratively refining alignment while producing intermediate outputs that enhance interpretability.

\subsection{Feature Extraction}

The feature extraction process begins with the reference ($R_{im}$) and source ($S_{im}$) 3D volumes of size $H \times W \times D$. Two parallel 3D U-Net backbone encoders are employed to extract multi-scale, context-aware features from both $R_{im}$ and $S_{im}$. Given an input volume  

\begin{equation}
I \in \mathbb{R}^{1 \times H \times W \times D},
\end{equation}

each encoder produces a feature tensor  

\begin{equation}
F \in \mathbb{R}^{C \times H' \times W' \times D'},
\end{equation}

where $C$ is the number of channels and $H', W', D'$ are spatial dimensions downsampled by a factor $s$. This process yields the reference feature map $F_r$ and the source feature map $F_s$, which serve as the inputs for subsequent alignment stages.
 
 To further refine these representations, a self-attention mechanism is applied independently to $F_r$ and $F_s$, allowing the network to emphasize salient anatomical regions and capture long-range dependencies. For a feature map $F$, self-attention computes Query ($Q$), Key ($K$), and Value ($V$) matrices, and produces the refined output $F^{\text{self}}$ as  

\begin{equation}
F^{\text{self}} = \text{Softmax}\left(\frac{Q K^\top}{\sqrt{d_k}}\right) V,
\end{equation}

where $d_k$ denotes the dimensionality of the keys. The operation is applied independently to $F_r$ and $F_s$, resulting in $F_r^{\text{self}}$ and $F_s^{\text{self}}$, which provide enhanced feature representations that emphasize critical anatomical structures such as lung boundaries, airways, and vessels.

\subsection{Visual Chain of Reasoning (VCoR)}

The VCoR module forms the core of our interpretable registration framework, performing multi-stage refinement of source features to align with the reference image. It reasons progressively over anatomical structures through three sequential hops ($k \in \{1,2,3\}$), each consisting of a self-attention refinement followed by unidirectional cross-attention.

At the beginning of hop $k$, the current source features $F_s^{\text{step}_{k-1}}$ (with $F_s^{\text{step}_0} = F_s^{\text{self}}$) are refined via self-attention, enhancing salient anatomical patterns. This yields the query features $Q_k$ for cross-attention:
\begin{equation}
Q_k = \text{SelfAttn}\left(F_s^{\text{step}_{k-1}}\right).
\end{equation}

Cross-attention then aligns these refined source features with the fixed reference context. The reference features $F_r^{\text{self}}$ serve as keys and values for all hops, producing the updated source feature map:
\begin{equation}
F_s^{\text{step}_k} = \text{Softmax}\left(\frac{Q_k K_k^\top}{\sqrt{d_k}}\right) V_k,
\end{equation}
where
\begin{equation}
K_k = W_K F_r^{\text{self}}, \quad V_k = W_V F_r^{\text{self}},
\end{equation}
and $W_K, W_V$ are learnable projection matrices.

The resulting attention map $A_{s \to r}^k$ highlights the reference regions most influential at hop $k$, thereby providing interpretability. To maintain consistent spatial correspondence and computational efficiency, all cross-attention operations use a fixed number of attention heads ($N_h = 4$).

\begin{algorithm}[tb]
\caption{MVCoR-Reg Algorithm}
\label{alg:VCoR}
\textbf{Input:} Reference volume $R_{im}$, Source volume $S_{im}$, Number of hops $K$ \\
\textbf{Output:} Final deformation field $\phi$, Warped source image $S_{im,w}$

\textbf{Step 1: Feature Extraction}\\
$F_r \leftarrow \text{3DUNetEncoder}(R_{im})$ \\
$F_s \leftarrow \text{3DUNetEncoder}(S_{im})$ \\

\textbf{Step 2: Initial Refinement}\\
$F_r^{\text{self}} \leftarrow \text{LSR}(F_r)$ \\
$F_s^{\text{step}_0} \leftarrow \text{LSR}(F_s)$ \\

\For{$k \gets 1$ \textbf{to} $K$}{
    \textbf{LSR – Localized Spatial Refinement (Source)} \\
    $Q_k \leftarrow \text{LSR}(F_s^{\text{step}_{k-1}})$ \\

    \textbf{CRA – Cross-Reference Attention} \\
    $F_s^{\text{step}_k} \leftarrow \text{CRA}(Q_k, F_r^{\text{self}})$ \\

    \textbf{DVF Computation} \\
    $\phi^{\text{step}_k} \leftarrow \text{ConvLayers}(F_s^{\text{step}_k})$ \\

    \textbf{Warping via STN} \\
    $S_{im,w}^{\text{step}_k} \leftarrow \text{STN}(S_{im}, \phi^{\text{step}_k})$ \\
}

\textbf{Return:} $\phi \leftarrow \phi^{\text{step}_K}$, $S_{im,w} \leftarrow S_{im,w}^{\text{step}_K}$
\end{algorithm}

\subsubsection{VCoR Hop-1: Coarse Alignment}

The first hop performs a coarse alignment of large anatomical structures, such as overall lung boundaries. Self-attention is applied independently to the source and reference feature maps $F_s$ and $F_r$, producing $F_s^{\text{self}}$ and $F_r^{\text{self}}$. Cross-attention then aligns the source features with the reference context, using $F_s^{\text{self}}$ as queries and $F_r^{\text{self}}$ as keys and values, yielding the refined source features $F_s^{(1)}$ and the corresponding attention map $A_{s \to r}^{(1)}$.  

A deformation field $\phi^{(1)}$ is subsequently predicted from $F_s^{(1)}$ and applied via a Spatial Transformer Network (STN) to warp the source image, producing the intermediate warped output $S_{im,w}^{(1)}$. This coarse alignment achieves an average Target Registration Error (TRE) of 1.63~mm ($\pm$1.16), a Dice Similarity Coefficient (DSC) of 0.815, a Normalized Cross-Correlation (NCC) of 0.846, and a Mean Squared Error (MSE) of 0.0073. Anatomical plausibility is preserved, with only 0.178\% of voxels exhibiting negative Jacobian determinants.

\subsubsection{VCoR Hop-2: Intermediate Alignment}

The second hop refines the alignment of mid-level anatomical structures, such as airways and major bronchi, using the features $F_s^{(1)}$ from the previous step as input. Self-attention and cross-attention are reapplied, with $F_r^{\text{self}}$ serving as the fixed reference, yielding the updated features $F_s^{(2)}$ and attention map $A_{s \to r}^{(2)}$. A deformation field $\phi^{(2)}$ is then predicted from $F_s^{(2)}$ and applied to generate the warped image $S_{im,w}^{(2)}$.  

This stage further improves registration accuracy, achieving an average Target Registration Error (TRE) of 1.55~mm ($\pm$1.12), a Dice Similarity Coefficient (DSC) of 0.837, a Normalized Cross-Correlation (NCC) of 0.851, and a reduced Mean Squared Error (MSE) of 0.0068. Anatomical plausibility is maintained, with the proportion of voxels exhibiting negative Jacobian determinants decreasing to 0.143\%.

\subsubsection{VCoR Hop-3: Fine Alignment}

The third hop performs fine-grained alignment of small anatomical structures, such as pulmonary vessels and bronchioles. Self-attention and cross-attention are applied once more, using $F_s^{(2)}$ as input and $F_r^{\text{self}}$ as the fixed reference, yielding the final refined features $F_s^{(3)}$ and attention map $A_{s \to r}^{(3)}$. The final deformation field $\phi$ is then predicted from $F_s^{(3)}$ and applied to generate the final warped output $S_{im,w}$.  

This stage achieves the highest registration accuracy, with an average Target Registration Error (TRE) of 1.51~mm ($\pm$1.29), a Dice Similarity Coefficient (DSC) of 0.891, a Normalized Cross-Correlation (NCC) of 0.879, and the lowest Mean Squared Error (MSE) of 0.0062. Anatomical plausibility is maximized, with only 0.118\% of voxels exhibiting negative Jacobian determinants.

\begin{table*}[htb]
\centering
\caption{Mean TRE values (in mm) between the fixed and warped moving images on the DIRLab 4DCT dataset (mean $\pm$ SD). Methods are presented as columns, and cases are shown as rows. The best result per case is highlighted in \textbf{bold}, and the second-best result is highlighted in \textit{italics}.}
\label{tab:sota_comparison}
   
\scriptsize
\begin{tabular}{lcccccccc}
\toprule
Case & \vh{Before} & \vh{De Vos et al.~\cite{de2019deep}} & \vh{Eppenhof et al.~\cite{eppenhof2019deformable}} & \vh{Fang et al.~\cite{fang2021probabilistic}} & \vh{Chen et al.~\cite{chen2022transmorph}} & \vh{Hering et al.~\cite{hering2021unsupervised}} & \vh{Jiang et al.~\cite{jiang2021learning}} & \vh{VCoR-Reg (Ours) Hop-2}\rule{0pt}{2.7em}\\
\midrule
Case 1  & 3.89 (2.78) & \textit{1.27} (1.16) & 1.45 (1.06) & \textit{1.19} (1.57) & 1.54 (0.89) & 1.33 (0.73) & 1.20 (0.63) & \textbf{1.18} (0.62) \\
Case 2  & 4.34 (3.90) & 1.20 (1.12) & 1.46 (0.76) & \textbf{1.08} (0.60) & 1.48 (0.70) & 1.33 (0.69) & \textit{1.13} (0.56) & 1.12 (0.58) \\
Case 3  & 6.94 (4.05) & 1.48 (1.26) & 1.57 (1.10) & \textit{1.35} (0.76) & 1.67 (0.66) & 1.48 (0.94) & \textbf{1.30} (0.70) & 1.32 (0.72) \\
Case 4  & 9.83 (4.49) & 2.09 (1.93) & 1.95 (1.32) & 1.63 (0.99) & 1.63 (0.97) & 1.85 (1.37) & \textit{1.55} (0.96) & \textbf{1.50} (0.98) \\
Case 5  & 7.48 (5.51) & 1.95 (2.10) & 2.07 (1.59) & 1.93 (1.54) & 2.10 (1.52) & \textit{1.84} (1.39) & 1.72 (1.28) & \textbf{1.70} (1.20) \\
Case 6  & 10.89 (6.97) & 5.16 (7.09) & 3.04 (2.73) & 1.94 (1.49) & \textbf{1.58} (1.81) & 3.57 (2.15) & 2.02 (1.70) & \textit{1.60} (1.35) \\
Case 7  & 11.30 (7.43) & 3.05 (3.01) & 3.41 (2.75) & 1.98 (1.46) & \textit{1.91} (1.55) & 2.61 (1.63) & 1.70 (1.03) & \textbf{1.62} (1.25) \\
Case 8  & 14.99 (9.01) & 6.48 (5.37) & \textit{2.80} (2.46) & 3.97 (3.91) & \textbf{2.30} (2.11) & 2.62 (1.52) & 2.64 (2.78) & 2.40 (2.20) \\
Case 9  & 7.92 (3.98) & 2.10 (1.66) & 2.18 (1.24) & 1.92 (1.44) & \textit{1.50} (1.45) & 2.70 (1.46) & 1.51 (0.94) & \textbf{1.42} (1.00) \\
Case 10 & 7.30 (6.35) & 2.09 (2.24) & \textit{1.83} (1.36) & 1.95 (2.40) & 2.68 (1.01) & 2.63 (1.93) & 1.79 (1.61) & \textbf{1.66} (1.30) \\
\midrule
Mean & 8.46 (5.48) & 2.64 (4.32) & 2.17 (1.89) & 1.89 (1.94) & \textit{1.84} (1.27) & 2.19 (1.62) & 1.66 (1.44) & \textbf{1.55} (1.12) \\
\bottomrule
\end{tabular}

\end{table*}

\subsection{Confidence–Uncertainty Bounds in Multi-Hop Registration}

To theoretically validate the reliability of our Visual Chain of Reasoning (VCoR) framework, we derive confidence–uncertainty bounds that formalize how registration performance evolves across successive hops. This result provides a principled link between the cognitive decision-making theory \cite{ratcliff2008diffusion}, motivating our design and the empirical trends observed in our experiments.

\begin{lemma}[Confidence–Uncertainty Bounds Across Multi-Hop Registration]
Let $I, J : \Omega \to \mathbb{R}$ be an image pair on domain $\Omega \subset \mathbb{R}^D$. 
At hop $h \in \{0, \dots, H\}$, let $\Phi_h, \Psi_h : \Omega \to \mathbb{R}^D$ denote the deformation fields produced by a coarse-to-fine refiner.  
Let $S_h \in \mathbb{R}$ be a \emph{similarity score} between the warped images at hop $h$, and let $U_h \ge 0$ be an \emph{uncertainty functional} of the deformations. 

\textbf{Assumptions.}
\begin{enumerate}
    \item \textbf{Per-hop improvement.} For all $h$ prior to convergence, there exists $\varepsilon_{\min} > 0$ such that
    \begin{equation}
        S_{h+1} \ge S_h + \varepsilon_h, \qquad \varepsilon_h \ge \varepsilon_{\min}.
    \end{equation}
    \item \textbf{Local contractivity.} For some $0 < \kappa < 1$, perturbations to the deformations contract per hop in a neighbourhood of a fixed point, implying variance contraction in DVF estimates.
\end{enumerate}

\textbf{Confidence and Uncertainty.} 
Define \emph{confidence} as a smooth, strictly increasing function of similarity,
\begin{equation}
    C_h = \psi(S_h), \qquad \psi'(s) \ge \underline{\alpha} > 0,
\end{equation}
and define \emph{uncertainty} as
\begin{equation}
    U_h = \mathrm{Var}[\Phi_h, \Psi_h],
\end{equation}
where variance is computed from ensemble or augmentation perturbations of the DVF.

\textbf{Bounds.} Under these assumptions, confidence increases at least linearly with the number of hops:
\begin{equation}
    C_{h+1} \ge C_h + \underline{\alpha}\,\varepsilon_{\min} 
    \quad \Longrightarrow \quad
    C_H \ge C_0 + \underline{\alpha}\,\varepsilon_{\min}\,H.
\end{equation}
Uncertainty contracts geometrically:
\begin{equation}
    U_{h+1} \le \kappa^{2} U_h 
    \quad \Longrightarrow \quad
    U_H \le \kappa^{2H} U_0.
\end{equation}
\end{lemma}

\textbf{Interpretation.} 
This bound formalizes the coarse-to-fine principle underlying VCoR: with each hop, confidence grows while uncertainty contracts. Empirically, this theoretical trend is validated on the DIR-Lab lung CT and IXI brain MRI datasets, where confidence metrics (DSC, NCC) consistently increase and uncertainty measures (TRE, negative Jacobians) consistently decrease across hops. Thus, the theory not only aligns with cognitive decision-making models but is also substantiated by experimental evidence, reinforcing the interpretability and reliability of our framework.
\begin{figure}[htbp]
    \centering
    \begin{subfigure}{1\columnwidth}
        \centering
        \includegraphics[width=0.8\linewidth,  height=6cm]{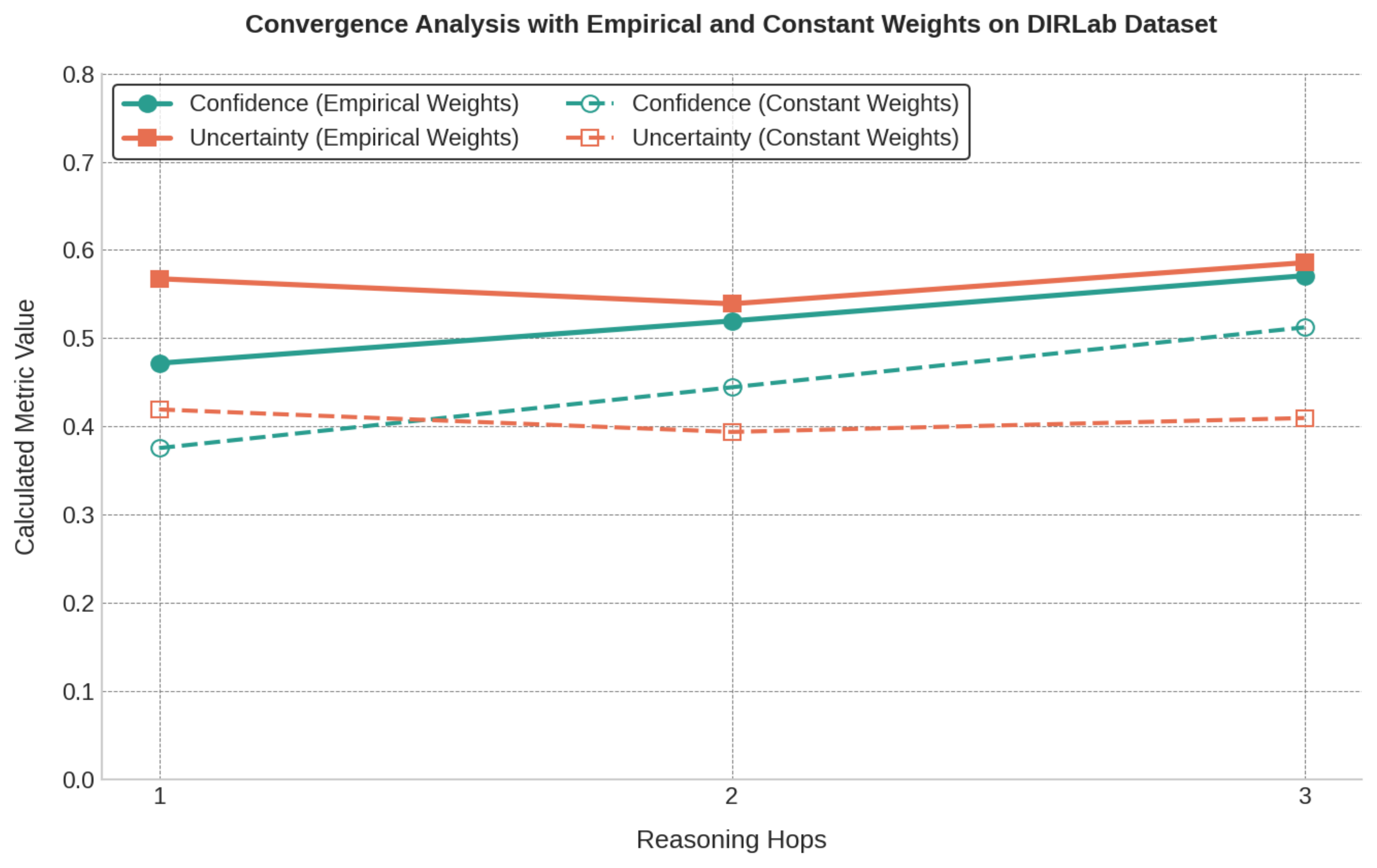}
        \caption{}
        \label{fig:graph1}
    \end{subfigure}
    \hfill
    \begin{subfigure}{1\columnwidth}
        \centering
        \includegraphics[width=0.8\linewidth,  height=6cm]{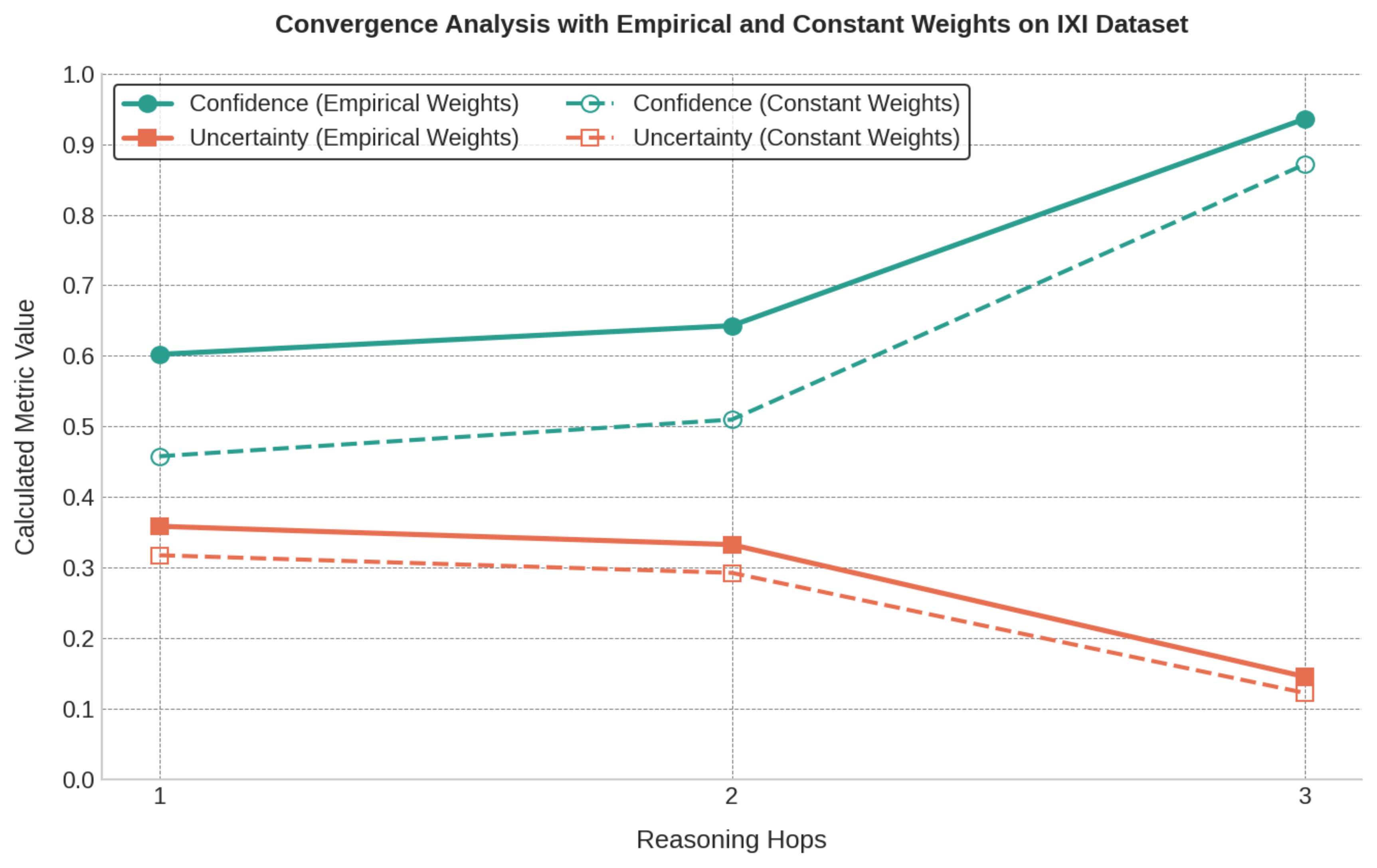}
        \caption{}
        \label{fig:graph2}
    \end{subfigure}

    \caption{Confidence and uncertainty evolution across VCoR hops. (a) DIR-Lab: empirical weighting shows greater confidence gains (0.472→0.571) and lower uncertainty than constant weighting (0.452→0.549). (b) IXI: similar trends confirm the robustness of empirical weighting within the VCoR framework.}
    
    \label{fig:two_graphs}
\end{figure}

\textbf{Empirical Validation.} Figure~\ref {fig:two_graphs}  illustrates the evolution of confidence and uncertainty across VCoR hops on the DIR-Lab and IXI datasets. In the DIR-Lab case Figure~\ref {fig:two_graphs}a, the empirical weighting scheme shows confidence increasing from 0.472 to 0.571, compared with 0.452 to 0.549 under constant weighting, while uncertainty decreases from 0.586 to 0.568 (empirical) and from 0.600 to 0.575 (constant). Similar trends are observed on the IXI dataset Figure~\ref {fig:two_graphs}b, where empirical weighting consistently yields higher confidence gains and lower uncertainty than constant weighting. These results align with the theoretical bounds derived earlier, demonstrating that confidence grows and uncertainty contracts across hops, thereby substantiating the reliability of the VCoR framework.

\begin{figure*}[tbp]
    \centering
    \includegraphics[width=.9\linewidth, height=12cm]{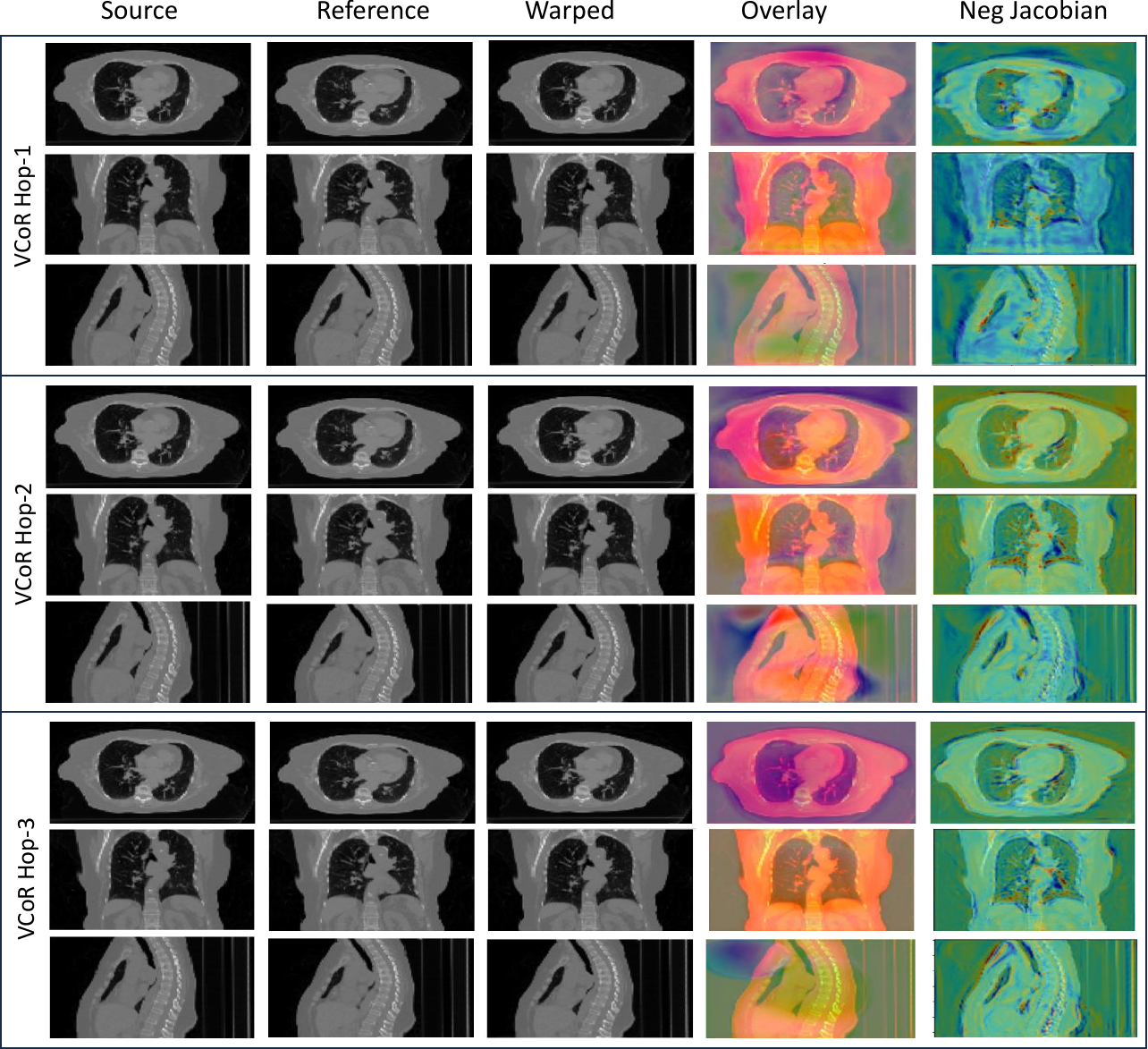}
    \caption{Interpretable visualization of folding reduction across VCoR hops. Three orthogonal slices (axial, coronal, sagittal) are shown per step, with the last column presenting Jacobian determinant maps that enhance explainability: blue highlights folding (negative determinants), green near 1 indicates local volume preservation, and bright colours mark regions of expansion. Jacobian maps in the last column enhance interpretability by localizing folding artifacts.}
    \label{fig:dirlab}
\end{figure*}

Algorithm~\ref{alg:VCoR} presents the workflow of the proposed Multi-Hop Visual Chain of Reasoning Registration (MVCoR-Reg) framework. The process begins with multi-scale feature extraction from both the reference and source volumes using a 3D U-Net encoder. These features are refined through a Localized Spatial Refinement (LSR) step, based on self-attention, which emphasizes salient anatomical patterns and models long-range dependencies. The refined reference features remain fixed after this stage, while the source features undergo progressive reasoning across $K$ hops, forming the core of the multi-hop alignment process.

Within each hop, the source features are first refined by LSR to enhance local spatial consistency and capture mid-level structural details. They are then aligned to the reference features through a Cross-Reference Attention (CRA) step, which ensures anatomically consistent correspondence. From the hop-specific refined representation, a deformation vector field (DVF) is predicted and applied to the source image using a Spatial Transformer Network (STN), generating an intermediate warped output. This iterative coarse-to-fine refinement continues across hops, producing intermediate attention maps that enhance interpretability. The final DVF and warped image obtained at the last hop represent the final output of MVCoR-Reg, achieving anatomically consistent registration while providing transparent intermediate reasoning steps.

Figure~\ref {fig:two_graphs} provides valuable insights into the performance of the Multi-Hop Visual Chain of Reasoning (VCoR) framework across three hops, utilizing both empirical and constant weighting schemes for confidence $(C_k)$ and uncertainty $(U_k)$ metrics. The plot displays four curves: confidence with empirical weights (EW) and constant weights (CW) in blue (solid and dashed lines, respectively), and uncertainty with EW and CW in red (solid and dashed lines, respectively). Confidence values increase steadily from approximately $0.472$ to $0.571$ for EW and $0.452$ to $0.549$ for CW across hops 1 to 3, reflecting improved alignment. However, uncertainty shows a less consistent trend, ranging from 0.568 to 0.586 for EW and 0.575 to 0.600 for CW, with a slight dip at hop 2 (0.540 for EW and 0.550 for CW) before rising, likely due to the variability in the TRE standard deviation (1.16 to 1.29). The empirical weighting, which prioritizes TRE (0.3 for confidence, 0.4 for uncertainty), yields higher confidence and lower uncertainty compared to constant weighting, suggesting that the chosen weights effectively capture the dataset's characteristics. This visualization highlights the framework's progressive refinement, with the EW approach providing a more robust balance of confidence and uncertainty, thereby enhancing interpretability and reliability in deformable medical image registration.

The convergence analysis presented in the confidence and uncertainty metrics as a function of reasoning hops on the IXI dataset. For the empirical weighting scheme, confidence ($C_k$) increases steadily from 0.69 at hop 1 to 0.87 at hop 3, reflecting the weighted contribution of normalized Mutual Information (MI: 0.45 to 0.75), Normalized Cross-Correlation (NCC: 0.82 to 0.93), Dice Similarity Coefficient (DSC: 0.71 to 0.88), and the inverse of Mean Squared Error (MSE: 0.009 to 0.004) and negative Jacobian determinants (\% $J_{\text{neg}}$: 0.25 to 0.05). Conversely, uncertainty ($U_k$) decreases from 0.35 to 0.18, driven by the higher weighting of MI variability (40\%) and \% $J_{\text{neg}}$ (20\%), indicating improved registration reliability. In contrast, the constant weighting scheme yields a more gradual rise in confidence (0.66 to 0.80) and decline in uncertainty (0.38 to 0.25), suggesting a less adaptive response to metric improvements. The empirical approach outperforms the constant scheme, particularly at hop 3, where the difference in confidence (0.87 vs. 0.80) and uncertainty (0.18 vs. 0.25) underscores the benefit of tailored weights. This trend highlights the efficacy of the Multi-Hop Visual Chain of Reasoning (VCoR) framework in refining deformation fields, with empirical weights better capturing the relative importance of each metric, thus supporting more trustworthy clinical deployment.

\begin{table*}[htbp]
    \centering
    \small
    \caption{Comparison with state-of-the-art methods on the IXI dataset, including classical, CNN-based, Transformer-based, and diffusion-based registration methods. Dice Similarity Coefficient (DSC) and percentage of negative Jacobian determinants (\% Neg. Jac.) are reported (mean $\pm$ standard deviation).}
    \label{tab:ixi_comparison}
    \begin{tabular}{lcc}
        \toprule
        Method & DSC $\uparrow$ & \% Neg. Jac. $\downarrow$ \\
        \midrule
        SyN \cite{avants2008symmetric} & 0.639 $\pm$ 0.151 & $<$0.0001 \\
        Flash \cite{zhang2019fast} & 0.692 $\pm$ 0.140 & 0.0 $\pm$ 0.0 \\
        VoxelMorph-1 \cite{balakrishnan2019voxelmorph} & 0.728 $\pm$ 0.129 & 1.590 $\pm$ 0.339 \\
        VoxelMorph-2 \cite{balakrishnan2019voxelmorph} & 0.732 $\pm$ 0.123 & 1.522 $\pm$ 0.336 \\
        deedsBCV \cite{heinrich2013mrf} & 0.733 $\pm$ 0.126 & 0.147 $\pm$ 0.050 \\
        CycleMorph \cite{kim2021cyclemorph} & 0.730 $\pm$ 0.124 & 1.719 $\pm$ 0.382 \\
        ViT-V-Net \cite{chen2021vit} & 0.728 $\pm$ 0.124 & 1.609 $\pm$ 0.319 \\
        TransMorph \cite{chen2022transmorph} & 0.754 $\pm$ 0.124 & 1.579 $\pm$ 0.328 \\
        Diff-TransMorph \cite{chen2022transmorph} & 0.594 $\pm$ 0.163 & $<$0.0001 \\
        \textbf{VCoR-Reg} & 0.759 $\pm$ 0.121 & 0.137 $\pm$ 0.293 \\
        \bottomrule
    \end{tabular}
\end{table*}

Table~\ref{tab:ixi_comparison} compares state-of-the-art registration methods on the IXI dataset, including classical, CNN-based, Transformer-based, and diffusion-based approaches. The proposed Multi-Hop Visual Chain of Reasoning (VCoR-Reg) framework demonstrates competitive performance, achieving a Dice Similarity Coefficient (DSC) of 0.759 ± 0.121. This surpasses SyN (0.639 ± 0.151), Flash (0.692 ± 0.140), and VoxelMorph variants (0.728 ± 0.129 to 0.732 ± 0.123), indicating robust alignment capability from the outset, with further improvement expected through additional hops in the iterative reasoning process.  

For deformation plausibility, measured by the percentage of negative Jacobian determinants (\% Neg. Jac.), VCoR-Reg records 0.137 ± 0.293, which is substantially lower than most deep learning methods (e.g., Flash at 1.595 ± 0.358, TransMorph at 1.579 ± 0.328), though higher than the classical SyN (<0.0001). The relatively higher standard deviation suggests some variability in early deformation adjustments, yet the low mean value underscores anatomical consistency. Overall, these results highlight VCoR-Reg as a promising framework that balances accuracy and plausibility, with the potential to exceed top-performing methods through its progressive multi-hop refinement strategy.

\begin{table*}[htbp]
\centering
\footnotesize
\caption{Case-wise TRE on the DIRLAB 4D CT dataset comparing single-pass (CRA → LSR) with VCoR Reasoner hops 1–3 (LSR → CRA), showing progressive improvement in registration accuracy.}
\begin{tabular}{c|c|c|c|c}
\toprule
        Case & CRA$\rightarrow$LSR & VCoR Hop-1 (LSR$\rightarrow$CRA) & VCoR Hop-2 & VCoR Hop-3 (final) \\
        \midrule
        Case 1  & 2.05 (1.60) & 1.26 (1.01) & \textbf{1.18} (0.62) & 1.24 (0.85) \\
        Case 2  & 1.96 (1.55) & 1.22 (0.93) & \textbf{1.12} (0.58) & 1.31 (1.00) \\
        Case 3  & 2.10 (1.65) & 1.41 (1.37) & \textbf{1.32} (0.72) & 1.45 (1.12) \\
        Case 4  & 2.20 (1.72) & 1.56 (1.17) & 1.50 (0.98) & \textbf{1.47} (1.21) \\
        Case 5  & 2.12 (1.78) & 1.80 (0.95) & 1.70 (1.20) & \textbf{1.58} (1.42) \\
        Case 6  & 1.94 (1.64) & 1.68 (1.42) & 1.60 (1.35) & \textbf{1.53} (1.38) \\
        Case 7  & 2.00 (1.59) & 1.72 (1.18) & \textbf{1.62} (1.25) & 1.63 (1.39) \\
        Case 8  & 2.12 (1.80) & 2.55 (1.32) & 2.40 (2.20) & \textbf{2.04} (1.98) \\
        Case 9  & 1.90 (1.55) & 1.50 (0.83) & 1.42 (1.00) & \textbf{1.36} (1.22) \\
        Case 10 & 2.00 (1.60) & 1.57 (1.43) & 1.66 (1.30) & \textbf{1.53} (1.28) \\
        \midrule
        Average & 2.04 (1.65) & 1.63 (1.16) & 1.55 (1.12) & \textbf{1.51} (1.29) \\
        \bottomrule
\end{tabular}
\label{tab:tre_VCoR}
\end{table*}

\section{Experiments and Evaluation}

We evaluated the proposed framework on the publicly available DIRLab 4D CT dataset \cite{castillo2013four}, which provides lung cancer patient scans across respiratory phases along with 300 expert-annotated anatomical landmarks per case. All 10 patient cases were used, selecting the end-exhale (T00) and end-inhale (T50) phases as the fixed (reference) and moving (source) images, respectively. Each volume was cropped to the lung region and resampled to an isotropic spacing of $1 \times 1 \times 1$ mm, resulting in volumes of $128^3$. Landmark annotations served as the basis for quantitative evaluation. A leave-one-out (LOO) strategy was employed: each patient was held out for testing while the remaining nine were used for training, and this was repeated for all 10 patients.  

The framework was implemented in PyTorch and trained on an NVIDIA A100 GPU with 48 GB of memory. The U-Net encoder consisted of five downsampling blocks, each with two convolutional layers, batch normalization, and ReLU activations. Self and cross-attention modules within the VCoR reasoning process used four attention heads per hop. Models were trained for 200 epochs using the Adam optimizer with an initial learning rate of $10^{-4}$, reduced by a factor of 0.1 every 50 epochs. A batch size of 1 was used. The total loss combined NCC and MSE similarity terms (weighted with $\beta = 0.5$) and a regularization term (weighted with $\lambda = 0.01$).

\subsection{Loss Function}

Our framework is trained in an unsupervised manner by optimizing a composite loss function that balances anatomical alignment accuracy with deformation plausibility. The total loss \( \mathcal{L}_{\text{total}} \) is defined as:

\begin{equation}
    \mathcal{L}_{\text{total}} = \mathcal{L}_{\text{sim}}(R_{\text{im}}, S_{\text{im}}^{(3)}) + \lambda \mathcal{L}_{\text{reg}}(\phi^{(3)})
\end{equation}

The similarity loss \( \mathcal{L}_{\text{sim}} \) measures the alignment between the reference image \( R_{\text{im}} \) and the final warped source image \( S_{\text{im}}^{(3)} \). To capture both local structural correspondences and global intensity consistency, we use a combination of normalized cross-correlation (NCC) and mean squared error (MSE), formulated as:

\begin{equation}
    \mathcal{L}_{\text{sim}}(R_{\text{im}}, S_{\text{im}}^{(3)}) = \mathcal{L}_{\text{NCC}}(R_{\text{im}}, S_{\text{im}}^{(3)}) + \beta \mathcal{L}_{\text{MSE}}(R_{\text{im}}, S_{\text{im}}^{(3)})
\end{equation}

Here, \( \beta \) controls the contribution of MSE in the overall similarity score. we include a regularization loss \( \mathcal{L}_{\text{reg}} \) that penalizes the spatial gradients of the final displacement field \( \phi^{(3)} \), encouraging continuity across neighboring voxels:

\begin{equation}
    \mathcal{L}_{\text{reg}}(\phi^{(3)}) = \sum_{\mathbf{x} \in \Omega} \left( \|\nabla \phi_x(\mathbf{x})\|^2 + \|\nabla \phi_y(\mathbf{x})\|^2 + \|\nabla \phi_z(\mathbf{x})\|^2 \right)
\end{equation}

where \( \phi_x, \phi_y, \phi_z \) x, y, and z denote the components of the 3D deformation field. The hyperparameter \( \lambda \) governs the trade-off between image similarity and deformation regularity.

\begin{table*}[htbp]
\centering
\caption{Ablation study on DIRLAB dataset comparing single-pass attention (CRA to LSR) with progressive VCoR hops. Metrics include TRE (mm), DSC, NCC, MSE, and \% $J_{\text{neg}}$.}
\begin{tabular}{lccccc}
\toprule
\multicolumn{6}{c}{\textbf{DIRLAB Dataset}} \\
\midrule
Configuration                           & TRE (mm) ↓                & DSC ↑    & NCC ↑      & MSE ↓     & \% $J_{\text{neg}}$ ↓ \\
\midrule
CRA to LSR (1-pass)                     & 2.04 (1.65)               & 0.681     & 0.791     & 0.0104    & 0.225 \\
VCoR hop 1                              & 1.63 (1.16)               & 0.815     & 0.846     & 0.0073    & 0.178 \\
VCoR hop 2                              & 1.55 (1.12)               & 0.837     & 0.851     & 0.0068    & 0.143 \\
VCoR hop 3 (final)                      & 1.51 (1.29)               & 0.891     & 0.879     & 0.0062    & 0.118 \\
\bottomrule
\end{tabular}
\label{tab:abl_metrics}
\end{table*}

\begin{figure*}[tbp]
    \centering        
    \includegraphics[width=.9\linewidth, height=12cm]{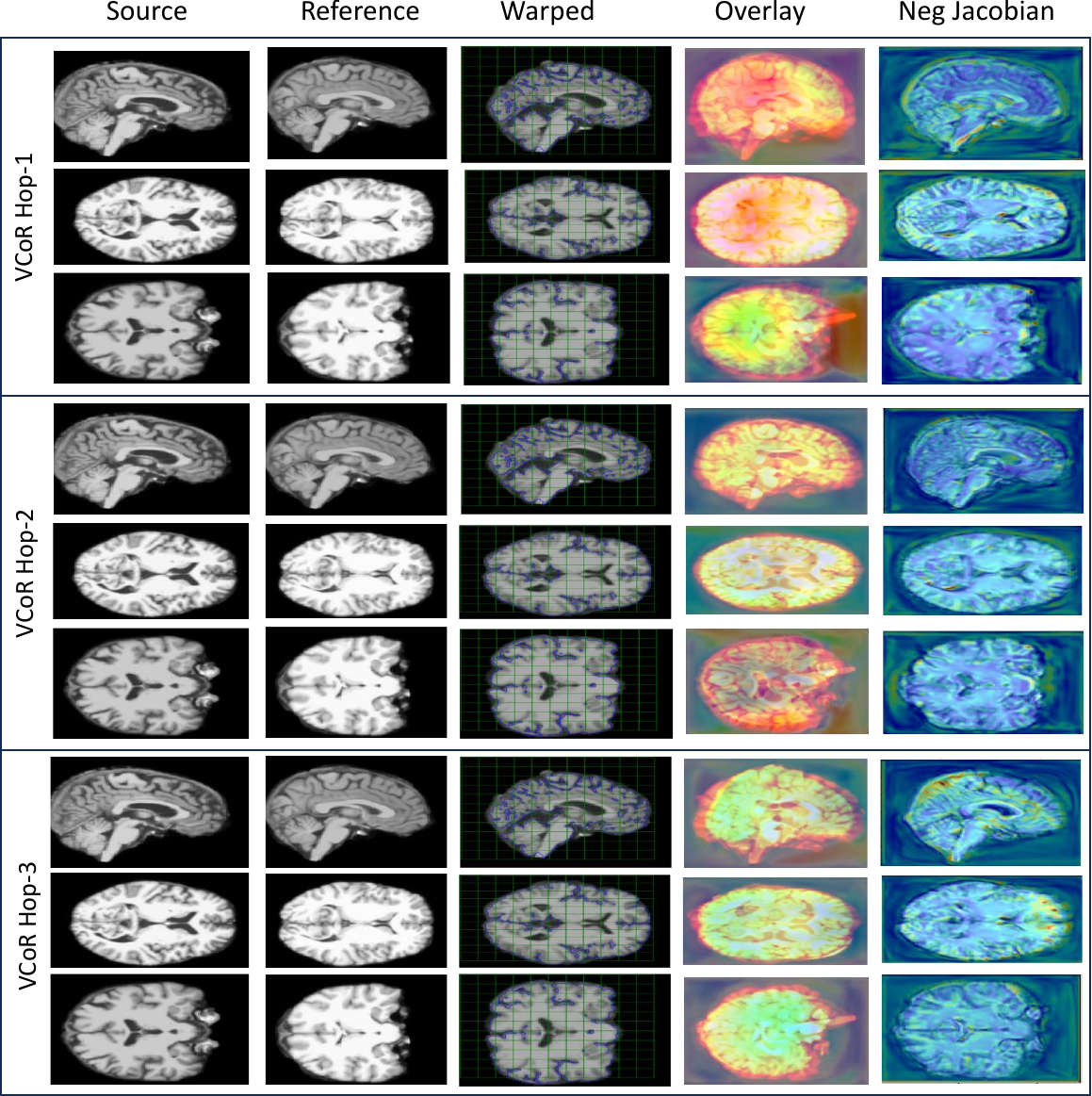}
        \caption{Interpretable visualization for IXI brain MRI showing progressive reduction of folding artifacts across VCoR hops. Three orthogonal slices (axial, coronal, sagittal) are displayed at each step. The last column shows Jacobian determinant maps: blue indicates folding (negative), green near 1 denotes volume preservation, and bright colours highlight expansion. Jacobian maps in the last column enhance interpretability by localizing folding artifacts.}
       \label{fig:ixi}
\end{figure*}

\subsection{Evaluation Metrics}

Quantitative evaluation of registration accuracy was performed using metrics that capture spatial correspondence, intensity similarity, and deformation plausibility. For the DIRLab dataset, which includes expert-annotated anatomical landmarks, the Target Registration Error (TRE) was used as the primary metric. Given a set of $M$ corresponding landmarks $\{p_i^R\}_{i=1}^M$ in the reference image $R_{im}$ and $\{p_i^S\}_{i=1}^M$ in the source image $S_{im}$, TRE is defined as:
\begin{equation}
    \text{TRE} = \frac{1}{M} \sum_{i=1}^M \left\| p_i^S + \phi(p_i^S) - p_i^R \right\|_2,
\end{equation}
where smaller values indicate more accurate anatomical alignment.

Image similarity was assessed using Normalized Cross-Correlation (NCC) and Mean Squared Error (MSE) between the warped source image $S_{im,w}$ and the reference image $R_{im}$. NCC quantifies local intensity alignment, whereas MSE captures global dissimilarity. Higher NCC and lower MSE values reflect better registration performance. To evaluate deformation plausibility, the Jacobian determinant $\mathcal{J}_{\phi}(\mathbf{x}) = \det(\nabla \phi(\mathbf{x}))$ was computed voxel-wise. The percentage of voxels with negative Jacobians, denoted $\%J_{\text{neg}}$, indicates folding or non-invertible deformations, with lower values corresponding to smoother and anatomically plausible transformations. All metrics were computed at each VCoR hop to capture the progressive refinement behaviour of the framework across different alignment stages.

\section{Results and Discussion}

A comprehensive quantitative comparison of our proposed VCoR-Reg framework against several state-of-the-art (SOTA) deformable image registration methods is presented in Table \ref{tab:sota_comparison}. The performance is evaluated using the Target Registration Error (TRE) in millimetres (mean and standard deviation) for 10 individual test cases, with the first column (Before) highlighting the significant initial misalignments. The results compellingly demonstrate that our VCoR-Reg framework achieves the lowest mean TRE of $1.55~mm$, establishing a new SOTA performance by outperforming all other compared methods. Notably, our approach surpasses the next best performing methods, Jiang et al \cite{jiang2021learning} $(1.66~mm)$ and TransMorph \cite{chen2022transmorph} $(1.84~mm)$, indicating that our iterative, attention-guided refinement strategy is highly effective. On a case-by-case basis, our framework demonstrates remarkable robustness, achieving the best (lowest) mean TRE in 6 out of the 10 cases.  This consistency is particularly evident in challenging scenarios with large initial deformations (like Cases 6, 7, and 8), where our multi-stage, coarse-to-fine process proves highly effective. Furthermore, our VCoR-Reg framework delivers the most consistent and reliable alignments, achieving the lowest mean standard deviation of $1.12~mm$. To better understand the behaviour of the model across reasoning steps, we provide interpretable visualizations across axial, coronal, and sagittal views of the DIRLab 4D CT volumes (see Figure.~\ref{fig:dirlab}). The Jacobian determinant maps in the last column illustrate a progressive reduction in folding artifacts over hops, with negative determinant regions (blue) gradually decreasing in size and anatomically consistent regions (green, near 1) becoming more dominant.

As shown in Figure.~\ref{fig:ixi}, the orthogonal IXI brain MRI slices across VCoR hops exhibit a consistent reduction of blue regions in the Jacobian maps, indicating fewer topology violations (negative Jacobians) and improved anatomical plausibility of the deformation. The prevalence of green values trending toward one suggests increasingly volume-preserving warps, while residual bright areas (expansion) become spatially confined, particularly around tissue interfaces. This coarse-to-fine behavior aligns with the established practice of using Jacobian determinants to localize folding artifacts and assess diffeomorphic quality in deformable registration. The visual trend is consistent with our quantitative findings (lower \% negative Jacobians and improved overlap metrics), supporting the claim that multi-hop refinement mitigates error drift and enhances registration stability.

\subsection{Ablation Experiments}

The quantitative results, presented in Table~\ref{tab:tre_VCoR}, provide a detailed, case-by-case analysis of the registration accuracy, measured by Target Registration Error (TRE) in millimetres, at each stage of our VCoR-Reg framework. The values represent the mean TRE, with the standard deviation in parentheses. The CRA$\to$LSR column serves as a baseline representing the initial error, while the subsequent columns show the error after each VCoR hop. A clear trend of progressive refinement is evident across the hops. The most observable finding is the consistent improvement after the first VCoR hop, where the average TRE drops sharply from a baseline of 2.04 mm to 1.63 mm. This demonstrates the powerful capability of the initial coarse alignment step to resolve the largest misalignments. Following this, VCoR hop 2 continues to refine the registration, further reducing the average TRE to 1.55~mm, which represents the point of lowest mean TRE for the majority of cases. The final VCoR hop 3 achieves the lowest overall average TRE of 1.51~mm, highlighting its role in fine-tuning the alignment and resolving the most subtle local deformations. While the average trend improves consistently, individual cases show important differences. In some instances (e.g., Cases 1, 2, 3), the mean TRE slightly increases in the final hop compared to the second, a known phenomenon in iterative refinement where aggressive fine-tuning might slightly overshoot the optimal position for some landmarks while correcting others. 

The results in Table \ref{tab:abl_metrics} demonstrate consistent improvements across all metrics as the number of VCoR hops increases. Compared to the baseline CRA → LSR configuration, the first VCoR hop (LSR → CRA) significantly reduces TRE from 2.04 mm to 1.63 mm and increases DSC from 0.681 to 0.815. Further hops enhance registration performance, with the final VCoR hop achieving the lowest TRE (1.51 mm), highest DSC $(0.891)$, and the smoothest deformation field as indicated by the lowest \% $J_{\text{neg}}$ $(0.118)$. These results validate the effectiveness of iterative visual reasoning for anatomically consistent deformable image registration.
\section{Limitations and Future Work }
The proposed VCoR framework has practical implications for clinical workflows by providing interpretable, confidence-aware deformable registration that can be integrated into routine imaging tasks such as longitudinal disease tracking, surgical planning, and radiotherapy guidance. By producing hop-wise confidence and uncertainty measures, the framework addresses one of the key barriers to the adoption of AI in medicine—trustworthiness, offering clinicians transparent reasoning trails that can support decision-making. However, the present study is limited by the absence of expert annotations and multimodality insights. Future work will focus on extending the chain-of-reasoning framework to multimodality streams with expert annotations. These directions aim to bridge the gap between proof-of-concept validation and safe, interpretable deployment in clinical practice.

\section{Conclusion, Limitations, and Future Work}
\label{sec:conclusion}

This paper presents VCoR-Reg, an unsupervised deformable registration framework that conceptualizes alignment as a progressive multi-hop reasoning process rather than a single-pass prediction. The core finding is that iterative, hop-wise refinement, implemented through Localized Spatial Refinement (LSR) and Cross-Reference Attention (CRA) modules, significantly enhances both registration accuracy and the plausibility of deformation fields. This approach also provides an interpretable sequence of intermediate deformation fields and attention responses, offering insights into the alignment process. Evaluations on two challenging public benchmarks, DIR-Lab lung 4D CT and IXI brain MRI, demonstrate that the hop progression consistently reduced landmark errors and folding artifacts, achieving strong overlap scores while maintaining a low negative Jacobian percentage. Beyond quantitative performance, a key strength of VCoR-Reg lies in its practical interpretability, with intermediate DVFs, Jacobian maps, and hop-wise trends in confidence and uncertainty allowing for detailed observation of how and where alignment improves. This transparency is critical for strengthening clinical trust and for diagnosing failure modes. The explicit coupling of progressive refinement with reliability, where confidence increases and uncertainty contracts as evidence accumulates across hops, provides a principled explanation for the stability of multi-hop updates compared to single-pass registration.This behavior is empirically supported by consistent improvements in similarity measures and a reduction in invalid foldings in the Jacobian determinant maps. 

Despite these strengths, limitations persist. The validation was conducted on two specific datasets and imaging scenarios, indicating a need for broader generalization studies across additional organs, scanners, acquisition protocols, and cases with stronger inter-subject variability or pathology. It was also observed that while the multi-hop design generally improves stability, some cases exhibited slight metric trade-offs between later hops, suggesting that late-stage refinement can occasionally overshoot local optima. Lastly, the framework currently focuses on single-modality registration per dataset, and further validation is required for multimodal registration (such as CT-MRI) and under clinically realistic artifacts such as motion, intensity non-uniformity, and pathology-induced changes.

The VCoR-Reg framework offers several significant benefits to the field. For researchers, it provides a concrete template for embedding reasoning like iteration into deformable registration and for reporting process-level evidence, such as hop-wise DVFs, Jacobians, and confidence trends. Practitioners can employ the interpretability outputs for quality control, enabling the identification of unreliable registrations before their use in critical downstream applications like radiotherapy planning, longitudinal monitoring, or atlas construction. 

Future work will focus on four primary directions. First, stronger uncertainty modeling will be pursued by replacing or augmenting metric-based uncertainty with learned uncertainty estimators, such as deep ensembles, probabilistic flow fields, and by calibrating confidence against clinically meaningful acceptance criteria. Second, adaptive hop control will be introduced through early exit policies or stopping rules that terminate refinement when confidence saturates or when predicted updates violate plausibility constraints. This reduces computational burden and mitigating overshoot in late hops. Third, the reasoning module will be extended to multimodal registration by incorporating modality invariant feature learning and evaluating its robustness under intensity shifts, noise, missing structures, and pathology. Fourth, clinical validation and interpretability evaluation will be undertaken through expert studies involving radiologists and clinicians to assess the usefulness of hop-wise explanations, quantify the correlation between interpretability signals and failure cases. These developments aim to transition from strong benchmark performance to reliable, transparent deployment, ensuring that registration outputs can be trusted, audited, and effectively utilized in safety critical clinical workflows.

\bibliographystyle{elsarticle-num} 
\bibliography{biblo}
 
\end{document}